\documentclass[letterpaper, 10 pt, conference]{ieeeconf}  

\IEEEoverridecommandlockouts                              

\usepackage{amsmath} 
\usepackage{amssymb}  
\usepackage{comment}
\usepackage{subfigure}

\title{\LARGE \bf
Visuo-Tactile Zero-Shot Object Recognition \\
with Vision-Language Model
}

\author{Shiori Ueda$^{1}$, Atsushi Hashimoto$^{2}$, Masashi Hamaya$^{2}$, Kazutoshi Tanaka$^{2}$, Hideo Saito$^{1}$
\thanks{This work was supported by JSPS KAKENHI Grant Number 21H04910, JST SPRING Grant Number JPMJSP2123 and Grant-in-Aid for JSPS Research Fellow Grant Number JP24KJ1962.}
\thanks{$^{1}$ S. Ueda and H. Saito are with Keio University, Yokohama 223-8522, Japan.
        {\tt\small shiori.ueda@keio.jp}}%
\thanks{$^{2}$A. Hashimoto,  M. Hamaya, and K. Tanaka are with OMRON SINIC X Corporation, Hongo 5-24-5, Bunkyo-ku, Tokyo, Japan
        {\tt\small atsushi.hashimoto@sinicx.com}}%
}

\usepackage{xcolor}

\usepackage{graphicx}
\newcommand{\fref}[1]{{Fig.~\ref{#1}}}
\newcommand{\tref}[1]{{Tab.~\ref{#1}}}

\usepackage{booktabs}
\usepackage{multirow}
\usepackage{mlmath}
\newcommand{\seqxt}{(\vx_t)}
\newcommand{\seqxbart}{(\bar{\vx}_t)}
\newcommand{\seqxhatt}{(\hat{\vx}_t)}
\newcommand{\fenc}{\mathit{Encoder}}
\newcommand{\fdec}{\mathit{Decoder}}
\newcommand{\fvq}{\mathit{VQ}}
\newcommand{\db}{D_{\rm tac2txt}}
\newcommand{\tactxt}{\mathit{description}}

\newcommand{\dtstopping}{\Delta t_{\rm stopping}}
\newcommand{\tstop}{t_{\rm stop}}

\newif\ifarxiv \newcommand{\arxiv}{\arxivtrue}
\arxiv

\ifarxiv
\usepackage{fancyhdr}
\usepackage{url}

\fi

\begin{document}

\ifarxiv
\twocolumn[
\noindent
© 2024 IEEE. Personal use of this material is permitted. Permission from IEEE must be obtained for all other uses, in any current or future media, including reprinting/republishing this material for advertising or promotional purposes, creating new collective works, for resale or redistribution to servers or lists, or reuse of any copyrighted component of this work in other works.\\
]
\thispagestyle{empty}
\pagenumbering{gobble}
\clearpage
\fi
\maketitle
\thispagestyle{empty}
\ifarxiv
\else
\pagestyle{empty}
\fi


\begin{abstract}

Tactile perception is vital, especially when distinguishing visually similar objects. We propose an approach to incorporate tactile data into a Vision-Language Model (VLM) for visuo-tactile zero-shot object recognition. Our approach leverages the zero-shot capability of VLMs to infer tactile properties from the names of tactilely similar objects. The proposed method translates tactile data into a textual description solely by annotating object names for each tactile sequence during training, making it adaptable to various contexts with low training costs. The proposed method was evaluated on the FoodReplica and Cube datasets, demonstrating its effectiveness in recognizing objects that are difficult to distinguish by vision alone.

\end{abstract}

\section{\uppercase{Introduction}}
\label{sec:introduction}

Tactile perception is essential in recognizing and interacting with objects for both humans and robots equally~\cite{dahiya_tactile_2010}.
While humans can visually estimate some of an object's properties like shape and material~\cite{fleming_visual_2014}, we rely on touch to perceive other properties, such as hardness and elasticity. Vision alone can be misleading, particularly when distinguishing objects that appear similar or have been shaped uniformly. This issue extends to identifying the internal conditions of objects, like being boiled or frozen.
Tactile data is often vital to complement visual data. 
On the other hand, its necessity is highly context-dependent. Hence, a method should be adaptable to each context with low training costs in visuo-tactile applications. 

We focus on the task of zero-shot object recognition using both visual and tactile data. 
Zero-shot object recognition involves recognizing objects that are not included in the training dataset.
This study considers leveraging the zero-shot ability of vision language models (VLMs) \cite{liu_llava_2023, yang_dawn_2023}. VLM is one of the extensions of large language models (LLMs) that accept images and texts as inputs.
Some studies report that LLMs acquire common sense \cite{kalakonda_action-gpt_2022, zhao_large_2023}. We expect that VLMs also possess common sense since they are developed based on LLMs. Such common sense would help estimate tactile properties from the similarity to a known object. 

Tactile sensing involves a complex mix of actions and sensors.
Zero-shot performance on various vision tasks has dramatically improved in the last few years \cite{zhang_vision_2024}. This is due to the development of VLMs pre-trained on large-scale datasets \cite{radford_learning_2021}. 
Image and text data have standardized formats, allowing for unified data processing. On the other hand, each new action requires new data collection for tactile sensing.
A greedy approach for incorporating tactile data into VLMs is fine-tuning VLMs for each new action or sensor. This method requires storing fine-tuned VLM models for each action and sensor, which would be intractable. Instead, we focus on bridging tactile information to a VLM through object names without fine-tuning them.

\begin{figure}[tb]
    \centering
    \includegraphics[width=\columnwidth]{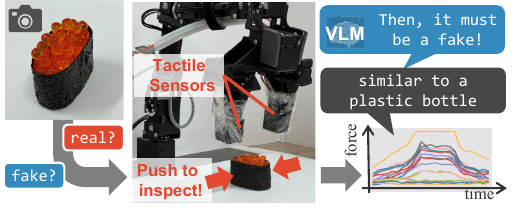}
    \caption{Task overview. We are addressing the problem of recognizing objects that are difficult to distinguish by vision alone. To facilitate recognition, the robot collects tactile information in addition to visual observation. Tactile signals are converted into a text description (e.g., "similar to \texttt{\$\{known reference object name(s)\}}") and fed to VLM. Based on the image, the textual tactile description, and common sense, VLM identifies the object class in a zero-shot manner.}
    \label{fig:typical_task}
\end{figure}

Based on these motivations, we propose a new approach for incorporating tactile data into the VLM for visuo-tactile zero-shot object recognition. \fref{fig:typical_task} illustrates the task overview. 
The proposed method utilizes a VLM solely by retrieving tactilely similar objects, which requires only annotating object names for each sample. 
Moreover, tactile similarities are learned without images or semantic labels, enabling the method to be easily replaced whenever the action or sensor changes.

We developed two visuo-tactile datasets, the FoodReplica dataset and the Cube dataset, for evaluation. They simulate the challenging scenario of recognizing objects by vision alone (real foods and replicas, and cube-shaped foods with different internal conditions). The evaluation on these datasets demonstrates that the proposed method outperforms the visual-only method.

Our contributions are as follows:
\begin{itemize}
    \item We propose a new approach incorporating tactile data into the VLM for visuo-tactile zero-shot object recognition. It leverages the common sense of VLMs to infer tactile properties from the names of tactilely similar objects.
    \item We introduce a tactile-to-text database that converts the tactile embedding into a textual description. It can be constructed with only tactile data and object names, facilitating adaptation to new actions or sensors at a low cost.
    \item We build two visuo-tactile datasets, FoodReplica and Cube, to simulate the challenging scenario of recognizing objects by vision alone.
\end{itemize}

\section{\uppercase{Related Works}}
\label{sec:related_works}

\subsection{Visuo-tactile Applications}

One of the common visuo-tactile applications is cross-modal learning \cite{gao_deep_2016, liu_visualtactile_2017, li_vito-transformer_2023}, which aims at extracting shared features and mapping them between different modalities. Some studies have exploited this property of cross-modal learning and applied it to zero-shot recognition \cite{liu_cross-modal_2020, fang_bidirectional_2024, yang_unitouch_2024}. Liu {\it et al.}~\cite{liu_cross-modal_2020} employed dictionary learning to transfer knowledge from visual to tactile data and utilized semantic labels to calculate similarity. Fang {\it et al.}~\cite{fang_bidirectional_2024} introduced a conditional flow module to bidirectionally map the latent spaces of visual and tactile data. Yang {\it et al.}~\cite{yang_unitouch_2024} aligned tactile embedding with a pre-trained image embedding to bind the tactile information to LLMs. These studies assume that the tactile sensation of an object can be estimated from its visual appearance. Our study differs from these studies in the sense that tactile data is used to obtain what cannot be estimated from visual data.

Visuo-tactile fusion is the other application, which aims to integrate different modalities to improve the performance of a task. Our study is grouped into this type of application.
In previous studies in object recognition, combining visual and tactile data has been shown to improve the recognition accuracy \cite{li_connecting_2019, takahashi_deep_2019, cai_visual-tactile_2021}. Visuo-tactile fusion has also been applied to zero-shot recognition \cite{abderrahmane_visuo-tactile_2018, fu_touch_2024}. Abderrahmane {\it et al.}~\cite{abderrahmane_visuo-tactile_2018} achieved zero-shot object recognition by predicting semantic labels from visual and tactile data. However, manually designed semantic labels are subjective and may not be consistent across different users. Experiments with multiple humans are needed to obtain a commonly recognized semantic label \cite{chu_robotic_2015, hassan_establishing_2023}, which is costly and limited to closed-set classes. Fu {\it et al.}~\cite{fu_touch_2024} utilized the VLM to generate semantic labels from visual and tactile data. It successfully aligns visual data, tactile data, and semantic labels, although the format of tactile data is fixed, and fine-tuning the VLM is required. 
Instead, we propose to bridge tactile data to a VLM via names of objects similar to the recognition target in tactile. The VLM infers the target's tactile properties based on their similarity to known objects, which requires no fine-tuning. In addition, the similarity is calculated independently from other modalities. Thus, this approach essentially assumes no specific data format.

\subsection{Actions for Tactile Perception}

Grasping is often used for tactile object recognition because it enables the collection of rich tactile data needed to identify the shapes and materials of objects from various positions and angles \cite{schmitz_tactile_2014, abderrahmane_haptic_2018}. 
The actions can be much simpler if the goal is solely to identify tactile properties like hardness or material type. 
Yuan {\it et al.}~\cite{yuan_shape-independent_2017} estimated the hardness of objects by pressing the target objects manually or by a robot hand in the normal direction. Several studies~\cite{chathuranga_robust_2015, baishya_robust_2016, taunyazov_towards_2019} recognized materials by sweeping sensors on the target objects. Yuan {\it et al.}~\cite{yuan_active_2018} predicted the material of the clothing by gripping the wrinkle of the clothing with a parallel gripper.
In this study, a pushing action is chosen rather than a grasping or sliding action to collect tactile data. It can be easily applied to various objects and can estimate hardness, which is difficult to infer by vision alone.

\subsection{Tactile Recognition of Foods}

Recognizing and handling fragile objects like foods is important for robotic applications \cite{WangFRA2022}.
Several studies focus on recognizing the tactile properties of foods \cite{gemici_learning_2014, sawhney_playing_2021}, but they often break the objects to measure these properties. Gemici {\it et al.} \cite{gemici_learning_2014} presented a method for obtaining the tactile properties of foods by interacting with them using forks and knives. Sawhney {\it et al.}~\cite{sawhney_playing_2021} constructed a dataset that includes the tactile properties of foods by squeezing them, pushing, and dropping their cut pieces. 
Yuan {\it et al.}~\cite{yuan_shape-independent_2017} estimated the hardness of objects, including foods (tomatoes), by pressing them without breaking them, but they do not provide a way to classify the foods. 
In this work, we aim to recognize the tactile properties of objects by softly pushing them with a parallel gripper to avoid breaking them, including fragile objects like macarons.

\section{\uppercase{Proposed Method}}
\label{sec:proposed_method}

\begin{figure*}[tb]
    \centering
    \includegraphics[width=\textwidth]
    {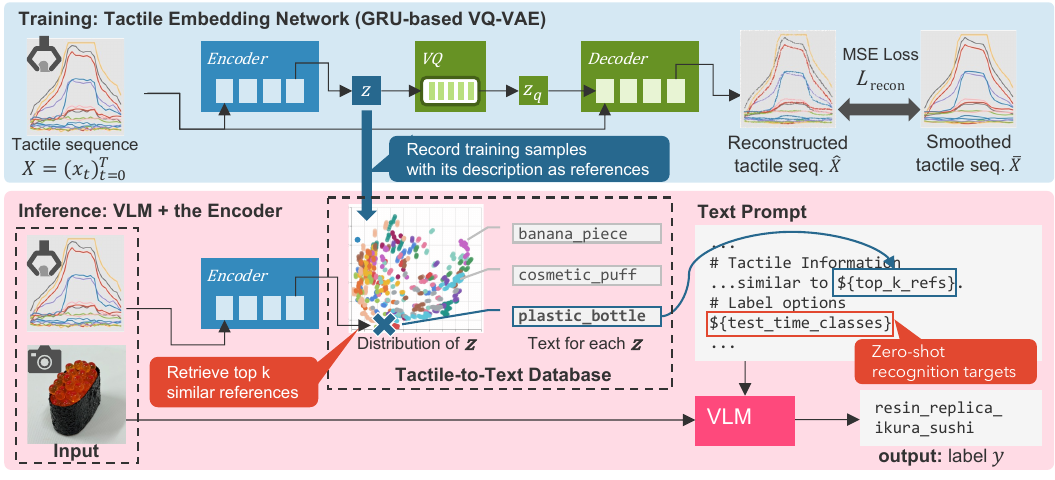}
    \caption{The proposed pipeline. The tactile embedding network learns the tactile embedding from the tactile sequence. These embeddings are converted to textual descriptions in the tactile-to-text database. During inference, the Vision Language Model (VLM) receives a textual description along with the visual image and outputs the most likely class label for the input object in a zero-shot manner.}
    \label{fig:pipeline}
\end{figure*}

\subsection{Problem Statement}
\label{sec:problem_statement}

Our task is zero-shot object recognition using visual and tactile data. Specifically, we aim to leverage tactile data as a cue to recognize objects that are difficult to distinguish using vision alone, as shown in \fref{fig:typical_task}.

Let $\vv$ be the input RGB image, and let $\mX=\seqxt_{t=1}^T$ be the input tactile sequence, where $\vx_t \in \sR^{d_{\rm x}}$ is the tactile signal at time $t$, and $T$ is the number of timesteps of the sequence. The output is a class label $y \in \fY_{\rm unseen}$, where $\fY_{\rm unseen}$ is the set of object classes given only at test time (and thus, inaccessible during the training phase).

\subsection{Tactile Embedding Network}
\label{sec:tactile_embedding_network}

As illustrated in \fref{fig:pipeline}, we extract a tactile embedding $\vz \in \sR^{d_{\rm z}}$ from the tactile sequence $\mX\in\sR^{T\times d_x}$ in an unsupervised manner.
We employ a sequence-to-sequence (Seq2Seq) model \cite{sutskever_sequence_2014} to learn the tactile embedding by reconstructing the input tactile sequence. Moreover, a vector quantization layer \cite{van_den_oord_neural_2017} is applied to convert the tactile embedding into a discrete latent variable. 
We construct a compact but discriminative model using a vector-quantization layer.

Let $\fenc$ be an RNN encoder, $\fvq$ be a vector quantization layer, and $\fdec$ be an RNN decoder. 
The tactile sequence $\mX$ is first input to $\fenc$. $\fenc$ outputs $\vz$, where $\vz$ is the hidden state of the RNN cell at the last timestep.
$\fvq$ quantizes the tactile embedding $\vz$ into a discrete tactile embedding $\vz_q$. In $\fvq$, the input $\vz$ is compared with a set of learnable codebook vectors $\mathcal{Z}_{\rm codebook} \in \sR^{K \times d_{\rm z}}$, where $K$ is the number of codebook vectors. The codebook vector $\vz \in \mathcal{Z}_{\rm codebook}$ that is closest to the input $\vz$ is selected as the quantized tactile embedding $\vz_q$.
The output of $\fvq$ is then used as the initial hidden state of $\fdec$.
With this initial hidden state $\vz_q$ and $\mX$, $\fdec$ decodes a sequence $\hat{X}=\seqxhatt_{t=1}^T$, which estimates a smoothed input $\bar{X}=\seqxbart_{t=1}^T$. Here, we apply a smoothing operation to prevent the model from overfitting to the noise.

The loss function $L$ follows VQ-VAE~\cite{van_den_oord_neural_2017} and is defined as:
\begin{equation}
    L = L_{\text{recon}} + L_{\text{vq}} + \beta L_{\text{commit}},
\end{equation}
where $L_{\text{recon}}$ is the reconstruction loss, $L_{\text{vq}}$ is the vector quantization loss, and $L_{\text{commit}}$ is the commitment loss.
$L_{\text{recon}}$ is defined as the mean squared error (MSE) between $\bar{\mX}$ and $\hat{\mX}$. $L_{\text{vq}}$ is the MSE between $\vz$ and $\vz_q$, where $\vz$ is detached from the computational graph. 
$L_{\text{commit}}$ is also the MSE between $\vz$ and $\vz_q$, but with detached $\vz_q$. $\beta$ is a hyperparameter that controls the strength of $L_{\text{commit}}$.

\subsection{Tactile-to-Text Database}
\label{sec:tactile-to-text_database}

The tactile embedding $\vz$ needs to be converted to a textual description to leverage the zero-shot ability of a VLM. To achieve this, we construct a tactile embedding database. The tactile-to-text database $\db : \mathcal{Z}_{\rm ref} \rightarrow \mathcal{Y}_{\rm ref}$ is a dict-type database that contains pairs of tactile embeddings and class labels, where $\mathcal{Z}_{\rm ref}$ is the set of tactile embeddings obtained from the TactileReference dataset.
The labels of the dataset are tied to textual descriptions.
To help the VLM better identify the tactile property of objects belonging to an unseen class $y\in \fY_{\rm unseen}$, we refer to top-$k$ nearest classes from $\fY_{\rm ref}$. Let $z$ be a tactile embedding obtained from an unseen class object at inference, and let $z_i\in \fZ_{\rm ref}$ be the reference tactile embeddings.
We retrieve the top-$k$ nearest classes with the L2-norm distance
$d(\vz, \vz_i) = \|\vz - \vz_i\|$.
We then concatenate textual descriptions for the top-$k$ classes with a comma to form $\tactxt$, which is included in the prompt for the VLM.

\subsection{Vision-Language Model as a Zero-shot Classifer}
\label{sec:vision-language_model_as_a_zero-shot_classifer}

\begin{figure}[tb]
    \centering
    \includegraphics[width=\columnwidth]{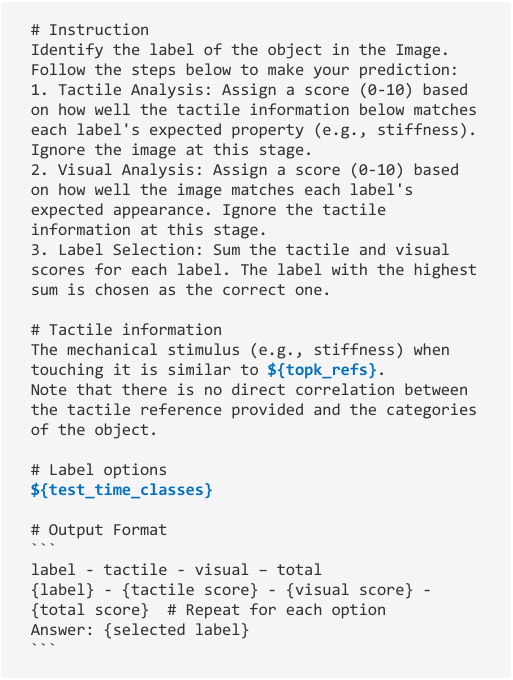}
    \caption{Prompt for visuo-tactile zero-shot object recognition. \texttt{\$\{topk\_refs\}} represents the reference classes of the top-$k$ nearest tactile embeddings. \texttt{\$\{test\_time\_classes\}} denotes the set of labels of the test dataset.}
    \label{fig:prompt}
\end{figure}

We utilize a VLM to perform zero-shot object recognition using vision and tactile textual data. 
We use GPT-4V \cite{yang_dawn_2023} as the VLM. The image $\vv$ and the textual tactile representation $\tactxt$ obtained in the previous subsection are input to the VLM. The prompt is designed to compel the VLM to evaluate visual and tactile likelihoods together. \fref{fig:prompt} shows the exact prompt used in this study. The prompt is a template with two slots: \texttt{topk\_refs} is replaced with $\tactxt$, and \texttt{test\_time\_classes} is replaced with $\fY_{\rm unseen}$.
The VLM selects a label $y$ from $\fY_{\rm unseen}$ based on $\vv$ and $\tactxt$ with this prompt.

\section{\uppercase{Experimental Setup}}
\label{sec:experimental_setup}

\subsection{Datasets}

We conducted experiments to evaluate the proposed method. We collected a tactile dataset, {\it TactileReference}, to train the tactile embedding network and construct the tactile-to-text database. In addition, we prepared two visuo-tactile datasets, {\it FoodReplica} and {\it Cube}, for testing.

\begin{figure*}[tb]
    \centering
    \includegraphics[width=\textwidth]{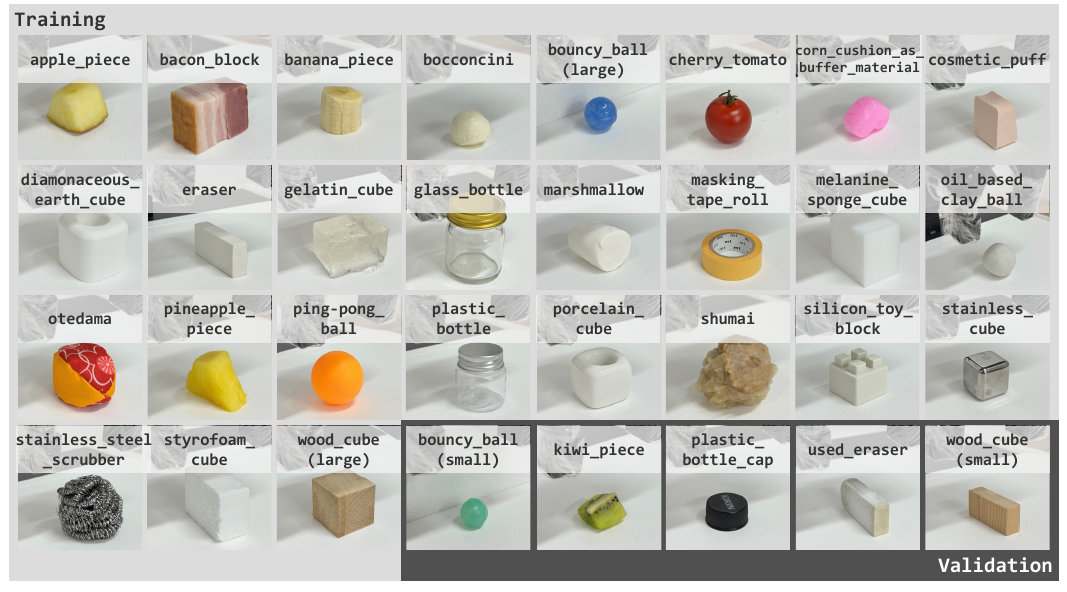}
    \caption{Snapshots of the TactileReference dataset. It is used in two modules. In the tactile embedding network module, the dataset is split into 27 classes for training and 5 classes for validation. In the tactile-to-text database module, all classes of the dataset are used to construct the tactile-to-text database.}
    \label{fig:train_dataset}
\end{figure*}

\fref{fig:train_dataset} shows the TactileReference dataset. 
It consists of 32 classes of labels $y \in \fY_{\rm ref}$ and the corresponding tactile sequences $\mX$. For network training, the dataset was split into 27 classes for training and 5 classes for validation. The validation set was used to monitor the convergence of the training and to determine the hyperparameters.  
We determined the text description corresponding to each label $y$ after verifying that GPT had not misinterpreted the object.
Note that we used only the training set to train the tactile embedding network but used both the training and validation sets for organizing the tactile-to-text database.

\begin{figure*}[tb]
    \begin{minipage}{1.0\textwidth}
    \centering
    \subfigure[the FoodReplica dataset]{%
        \includegraphics[width=\textwidth]{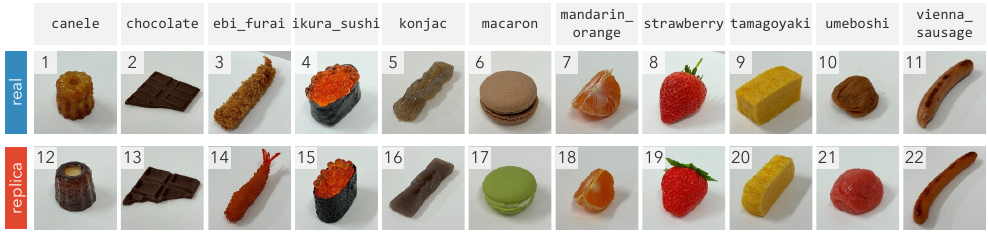}
    }\hfill%
    \end{minipage}\\
    \begin{minipage}{1.0\textwidth}
    \centering
    \subfigure[vision only]{%
        \includegraphics[width=0.48\columnwidth]{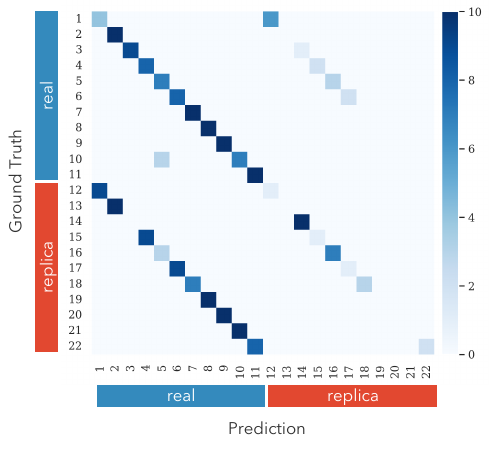}
    }\hfill%
    \subfigure[vision + tactile]{%
        \includegraphics[width=0.48\columnwidth]{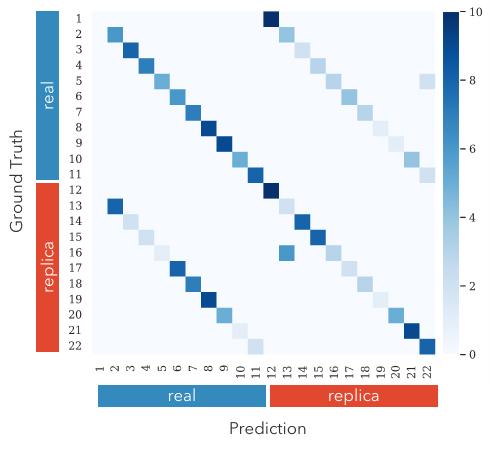}
    }
    \caption{The FoodReplica dataset and the results. Class labels of replicas are given in the \texttt{resin\_replica\_\$\{name\}} format. The vision-only method predicted replicas as real in most cases, while the proposed (vision + tactile) method achieved a balanced performance.}
    \label{fig:result_replica}
    \end{minipage}
\end{figure*}

The FoodReplica dataset, shown in \fref{fig:result_replica}~(a), was used to evaluate the performance of the proposed method. A dataset sample is given as a triplet $(\mX, \vv, y)$.
Food replicas are wax or plastic models of food. They are visually similar to real food but have different tactile properties. We chose food replicas as the target objects because they are among the most difficult objects to distinguish using vision alone.
It consists of 22 classes (11 pairs of real foods and their replicas) of labels $y \in \fY_{\rm unseen}$, visual images $\vv$, and tactile sequences $\mX$. 

\begin{figure}[tb]
    \centering
    \includegraphics[width=\columnwidth]{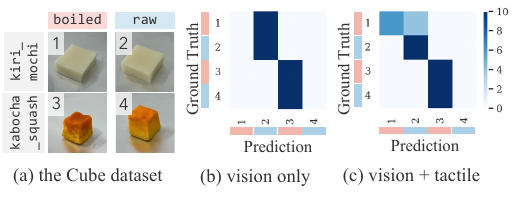}
    \caption{The Cube dataset and the results. Class labels are given in the \texttt{\$\{state\}\_\$\{name\}} format (e.g., \texttt{boiled\_kiri\_mochi}). Tactile data helped to distinguish raw and boiled kiri\_mochi, while kabocha\_squash remained difficult to distinguish even with the proposed method.}
    \label{fig:result_cube}
\end{figure}

The Cube dataset, shown in \fref{fig:result_cube}~(a), was also used to evaluate the performance of the proposed method, with samples also given as $(\mX, \vv, y)$.
It consists of 4 classes (2 objects $\times$ 2 conditions) of labels $y \in \fY_{\rm unseen}$, visual images $\vv$, and tactile sequences $\seqxt$.
It simulates a cooking situation, where objects are processed to have unified shapes but different internal conditions. We used kabocha\_squash and kiri\_mochi (a type of rice cake) as the target objects. We prepared two conditions for each object: one is a raw condition and the other is a boiled condition.
Both objects become soft when boiled while preserving their visual appearance.
We prepared this dataset to simulate a more practical context than FoodReplica.

\subsection{Hardware Settings}

The middle of \fref{fig:typical_task} shows our hardware settings.
We used a robot arm (OpenMANIPULATOR-X, RM-X52-TNM, Robotis Co., Ltd.) \cite{open_manipulator} with a parallel gripper. The distributed 3-axis tactile sensor (uSkin XR 1944, XELA Robotics
Co., Ltd.) \cite{u_skin} was attached to each of the two fingers of the gripper. The tactile sensor has a four-by-four taxel array. With these settings, we can obtain $d_{\rm x}=4 \times 4 \times 3 \times 2=96$-dimensional tactile signals at each timestep.

For the image input of the test datasets, we used an iPhone 15 Pro rear camera (77 mm, ƒ/2.8). During image capture, the target object was placed on a white cutting board for the FoodReplica dataset to simplify the background. For the Cube dataset, the target object was placed on an aluminum foil to ensure that the boundary between the white object and the background was visually distinguishable.

\subsection{Data Collection}

\begin{figure}[tb]
    \centering
    \includegraphics[width=\columnwidth]{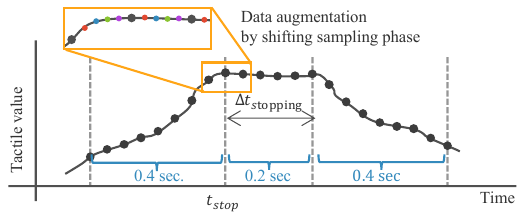}
    \caption{Visualization of the data collection process. We crop the sequence based on the timing of $\tstop$, with simple data augmentation by shifting the offset for subsampling. This cropping operation results in $\mX$ with $T=25$. }
    \label{fig:data_collection}
\end{figure}

Tactile sequences were obtained by softly pushing the target object with the gripper following predefined motions. The motions were as follows:
\begin{enumerate}
    \item Open the gripper above the target object.
    \item Lower the gripper to the position where the bottom of it is almost touching the surface of the table.
    \item Close the gripper until the norm of the tactile signal exceeds a threshold $x_{\rm th}$.
    \item Stop the gripper for $\dtstopping$. Record the timestamp $\tstop$ when the gripper stops. It is used for preprocessing the sequence.
    \item Open the gripper and go back to the initial position.
\end{enumerate}
We set $x_{\rm th}=0.004$ \footnote{The sensors were not calibrated and the value is sensor specific. We set $x_{\rm th}=0.003$ for gelatins exceptionally because the objects are far more fragile than others.} as a sufficiently gentle value and $\dtstopping=0.2$ s as a time long enough to observe delayed physical reactions, we repeated the process ten times for each label $y$. The target object was placed below the gripper and slightly moved for each process to make the tactile sequence different.

An ideal sequence $\mX$ starts when the gripper contacts the object and ends when the object is released. Unfortunately, it is difficult to detect such points precisely for versatile objects.
Instead, we crop the sequence centered on the most vital contact: the timing of stopping the gripper, as shown in \fref{fig:data_collection}. 
$\tstop$ is the timing when the gripper stops its motion with the strongest force. We recorded the sequence from 0.4 s before the $\tstop$ (during the gripper closing motion) to 0.6 s after the $\tstop$ (0.2 s for $\dtstopping$ and 0.4 s during the gripper opening motion), which is a total of 1.0 s.

We implemented the above with a simple data augmentation by subsampling. We observed the tactile data with a sampling rate of 125 Hz. Then, we subsampled the sequence at 25 Hz. By shifting the offset, we can obtain $5$ different sequences from one process in this setting. Overall, based on the timing of $\tstop$, we cropped the sequence with $T=25$ with $5$ augmented samples. 

In addition to the above subsampling, we obtained $\bar{\mX}$ by applying a Gaussian filter to the observed sequence with its standard deviation of $\sigma=3.0$.
To summarize, the number of samples for each $y$ was 50 (10 processes $\times$ 5 augmentations). This resulted in a total of 1350 samples (27 classes) for training and 250 (5 classes) for validation sets of TactileReference. For the other two test datasets (Food Replica and Cube), we collected 1 sample for each process, resulting in 10 samples from 10 processes for each $y\in \fY_{\rm unseen}$.

\subsection{Parameters for each module}

For $\fenc$, we used a 2-layer bi-directional GRU with a hidden size of 32, followed by a linear layer with an output size of 4. For $\fdec$, we used a 2-layer GRU with a hidden size of 32. For $\fvq$, we set $d_{\rm z}$ to 4 and the number of codebook vectors $K$ to 32. The output of $\fvq$ was input to a linear layer with an output size of 32, which was then set as the initial hidden state of the decoder $\fdec$. 
In the loss function, we set $\beta$ to 0.25.
We implemented our network using PyTorch 2.0.0 and trained it on a single NVIDIA GeForce RTX 3080. 
We used the Adam optimizer with a learning rate of 0.01. The batch size was 256. We trained the network until the validation loss converged (22 epochs).

During inference, we set $k=3$ and used the top 3 class labels to construct the tactile text $\tactxt$. We set the VLM's temperature to 0.0 to make the output deterministic.

\subsection{Evaluation Metrics}

We used accuracy as the evaluation metric, which is defined as the ratio of correctly recognized samples to the total number of test samples. We also used the confusion matrix to visualize the failure tendencies.

\section{\uppercase{Results}}
\label{sec:results}

We evaluated the proposed method on the FoodReplica and Cube datasets.
We also provide two ablation studies, one with a different choice of VLMs and one with a different size of the TactileReference dataset.

\subsection{Comparison with the Vision-only Baseline}

\begin{figure}[tb]
    \centering
    \includegraphics[width=\columnwidth]{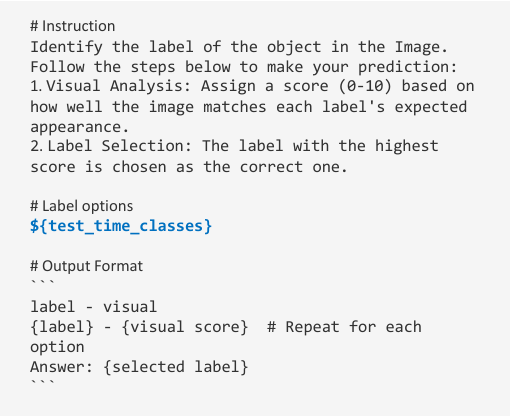}
    \caption{Prompt for vision-only baselines. The prompt is the same as \fref{fig:prompt} except that expressions related to the tactile data are removed.}    
    \label{fig:prompt_baseline}
\end{figure}

We compared the proposed method with the baseline, vision-only recognition using GPT-4V. It takes the image $\vv$ as input. For the baseline, we removed instructions related to tactile data from the proposed prompt, as shown in \fref{fig:prompt_baseline}.
In \tref{tab:result}, the first row (Vision-only~(1)) shows the accuracy of the baseline, and the last row (Our full model) shows the proposed method. The proposed method outperformed the baseline in both datasets.

\begin{table}[tb]
    \setlength\tabcolsep{4pt}
    \caption{The classification accuracy on FoodReplica and Cube.}
    \label{tab:result}
    \centering
    \begin{tabular}{lccccc}
        \toprule
        \multicolumn{4}{c}{Method} & \multicolumn{2}{c}{Accuracy (\%)} \\
        \multicolumn{1}{c}{Name} & Tactile & top-$k$ & VLM & FoodRep. & Cube \\
        \midrule
        Vision-only~(1) &-- & -- & GPT-4V & 53.6 & 50.0 \\
        Vision-only~(2) &-- & -- & LLaVA & 39.0 & 25.0 \\
        \midrule
        LLaVA variant & VQ-VAE & 3 & LLaVA & 39.1 & 25.0 \\
        $\beta$-VAE variant & $\beta$-VAE & 3 & GPT-4V & 52.5 & 50.0 \\
        top-1 variant & VQ-VAE & 1 & GPT-4V & \underline{58.4} & \underline{52.5} \\
        Our full model & VQ-VAE & 3 & GPT-4V & {\bf 58.9} & {\bf 65.0} \\
        \bottomrule
    \end{tabular}
\end{table}

\fref{fig:result_replica}~(b) shows the tendency of failure caused by Vision-only~(1), which tends to recognize any objects as real foods despite the given set of $\fY_{\rm unseen}$ in the prompt. On the other hand, as shown in \fref{fig:result_replica}~(c), the proposed method increases the chance of recognizing replicas as replicas.
This observation indicates that the proposed method can leverage the tactile data to complement the vision data.
While overall performance is improved by using both vision and tactile sensors, some regression is observed in cases such as completely misclassifying a real canele as a replica canele and confusing a replica konjac with a replica chocolate. The misclassification of the canele could be due to the real canele being as hard as the replica, resulting in similar tactile properties. Similarly, the replica konjac and the replica chocolate may have been confused because they are tactilely similar when pressed from the side, and their appearances were also similar.

\fref{fig:result_cube}~(b) and (c) show the confusion matrices of the baseline and the proposed method on the Cube dataset.
Both methods struggled to classify kabocha squash correctly, while the proposed method improved the accuracy for boiled kiri mochi.
Although the visual appearance of the boiled kiri mochi is quite similar to the raw one, the tactile properties are clearly different. As a result, many of the $\tactxt$ of the boiled kiri mochi were like "otedama" and "kiwi\_piece," which are soft objects, while those of the raw kiri mochi were like "wood\_cube" and "stainless\_cube," which are hard objects. It seems that these tactile textual descriptions helped the VLM to recognize the boiled kiri mochi correctly.
Meanwhile, even the proposed method completely misclassified the raw kabocha squash as the boiled one.
This could be because the tactile signals of the raw and boiled kabocha squash were similar and our method could not distinguish them. In this study, we defined the movement of the gripper as a soft touch to prevent the objects from breaking. However, the pressure was not enough to distinguish the raw and boiled kabocha squash. In future work, we need to consider the motion of the gripper to obtain more distinguishable tactile signals.

\subsection{Comparison with the variants of the proposed method}

We compared the proposed method with its variants, as shown in \tref{tab:result}.
The first variant is the baseline method with GPT-4V replaced by LLaVA \cite{liu_llava_2023} (Vision-only~(2)) and the second variant is the proposed method, which also uses LLaVA (LLaVA variant).
The third variant is the proposed method incorporating the reparameterization trick of the $\beta$-VAE \cite{higgins_beta-vae_2016} instead of the vector quantization layer ($\beta$-VAE variant).
The fourth variant is the proposed method but using only the top-1 nearest neighbor to organize $\tactxt$.

\tref{tab:result} shows that GPT-4V significantly outperforms LLaVA in this task. In addition, LLaVA hardly responded to tactile information in the prompt, which resulted in no performance increase. In contrast, our method improved the performance well with GPT-4V under the same conditions.
For tactile encoding, VQ-VAE (our full model) outperformed the $\beta$-VAE variant. Moreover, the $\beta$-VAE variant performed worse than the baseline. This indicates that the vector quantization layer constructs a better tactile embedding space that improves classification accuracy compared to the reparameterization trick of the $\beta$-VAE. 
Finally, the top-3 implementation (our full model) outperformed the top-1 variant. This indicates that three references could better identify the property of unseen objects. From the aspect of absolute accuracy, we confirmed that there is still a large room for improvement in this challenging task.

\subsection{Analysis of the number of reference classes}

\begin{figure}[tb]
    \centering
    \includegraphics[width=0.9\columnwidth]{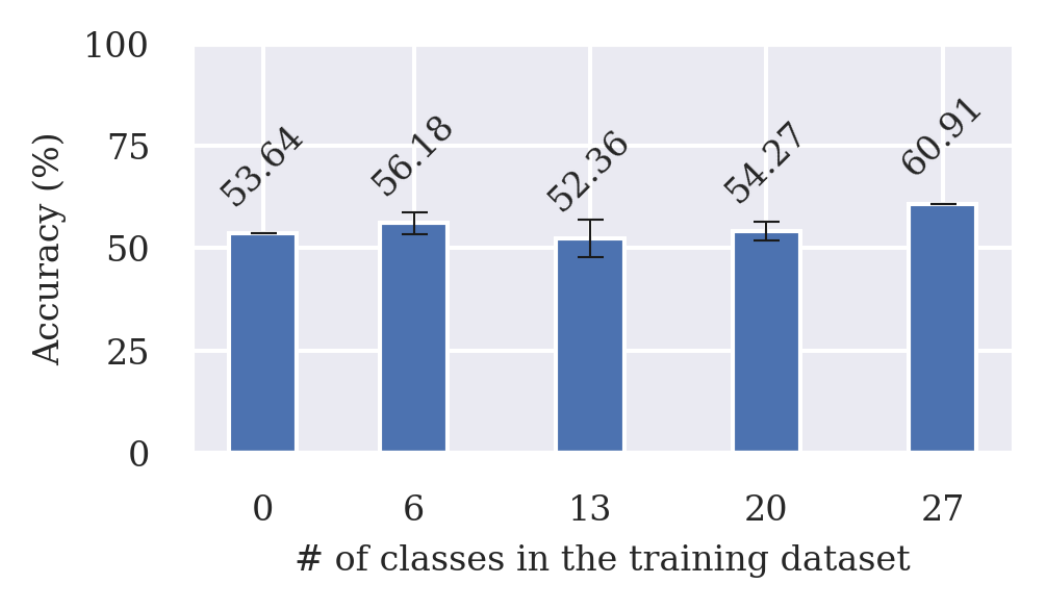}
    \caption{Accuracy of the proposed method with a controlled number of reference classes on the FoodReplica dataset.
    For the vision-only case and the full case, we tested the model once. For the other cases, we tested the model five times with different random combinations of training datasets and references. The plot shows the mean as a label and the standard deviation as an error bar for each case.}
    \label{fig:result_train_size}
\end{figure}

\fref{fig:result_train_size} analyzes the impact of the number of reference classes in the training dataset and tactile-to-text database on the recognition performance.
We obtained the results with $|\fY_{\rm ref}|=0$ (vision-only), 6, 13, 20, and 27 (full) on the FoodReplica dataset.
Here, we did not include the validation data in the tactile-to-text database to control the parameter systematically.

We repeated the experiment five times with different combinations of training datasets and references, except for the vision-only case ($|\fY_{\rm ref}|=0$) and the full case ($|\fY_{\rm ref}|=27$).
We could not find a clear trend in the accuracy, but the proposed method with the full training dataset achieved the highest accuracy. It indicates that the proposed method can leverage the tactile data more effectively with the full training dataset.

\section{\uppercase{Conclusions}}
\label{sec:conclusions}

We proposed a method for visuo-tactile zero-shot object recognition leveraging the common sense of VLMs. We constructed a tactile embedding network to extract a tactile embedding from the tactile sequence and a tactile-to-text database to convert the tactile embedding to a textual description. These were constructed from only the tactile sequences and class labels, eliminating the need for manual semantic labels and visual images bound to the tactile data.
We used GPT-4V as the VLM to perform zero-shot object recognition using vision and tactile textual data. We evaluated the proposed method on the FoodReplica and Cube datasets and compared it with the vision-only baseline and its variants. The proposed method outperformed the baseline in both datasets. 
Future work includes exploring the motion of the gripper to obtain more distinguishable tactile signals and considering the best practices for prompt design to incorporate the tactile data more effectively into various VLMs.






\section*{ACKNOWLEDGMENT}

The authors thank Reina Ishikawa for her invaluable support with robot operations.


\bibliographystyle{IEEEtran}
\bibliography{ref}

\end{document}